\documentclass[11pt,a4paper]{article}
\usepackage[hyperref]{naaclhlt2018}
\usepackage{times}
\usepackage{latexsym}
\usepackage{graphicx}
\usepackage{multirow}

\usepackage{url}

\aclfinalcopy 

\title{Contextualize, Show and Tell: A Neural Visual Storyteller}

\author{Diana Gonzalez-Rico \and  Gibran Fuentes-Pineda \\
    {\tt dianaglzrico@gmail.com} \and {\tt gibranfp@unam.mx} \\\\ Institute for Research in Applied Mathematics and Systems (IIMAS)\\ Universidad Nacional Aut\'onoma de M\'exico (UNAM)}

\date{}

\begin{document}
\maketitle
\begin{abstract}
  In this paper, we present a neural model for generating short stories from image sequences, which extends the image description model by Vinyals et al.~\cite{vinyals}. 
  This extension relies on an encoder LSTM to compute a context vector of each story from the image sequence. 
  This context vector is used as the first state of multiple independent decoder LSTMs, each of which generates the portion of the story corresponding to each image in the sequence by taking the image embedding as the first input. 
  Our model showed competitive results with the METEOR metric and human ratings in the internal track of the Visual Storytelling Challenge 2018.
\end{abstract}

\section{Introduction}
\begin{figure}[tb]
    \centering
    \includegraphics[trim={0 .2cm 0 0},clip,width=.5\textwidth]{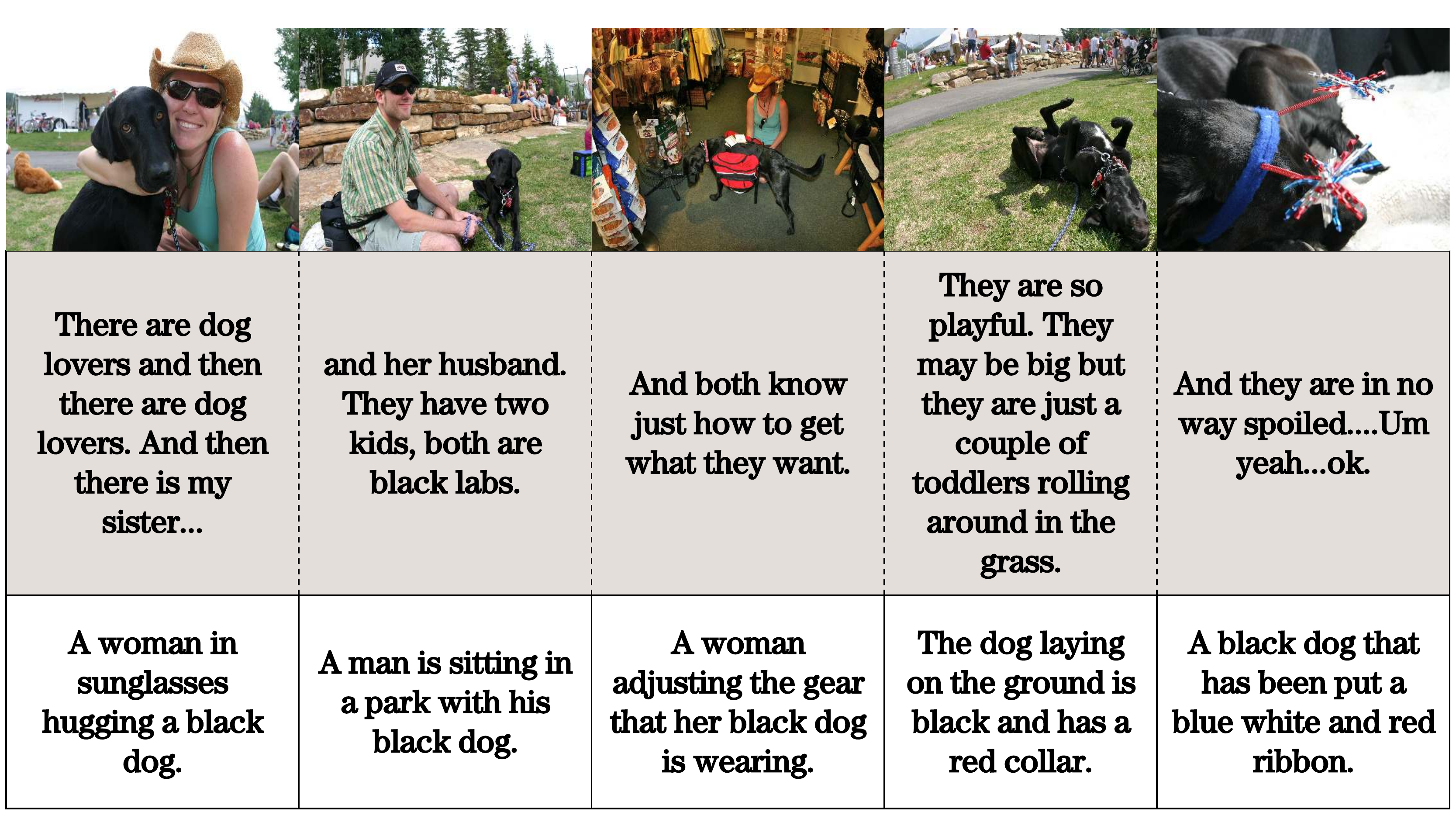}
    \caption{Examples of stories for images in sequence (above) and image descriptions in isolation (below) from the VIST dataset~\cite{huang2016}.}
    \label{fig:sisvsdii}
\end{figure}

Over the past few years, generating text from images and videos has gained a lot of attention in the Computer Vision and Natural Language Processing communities and several related tasks have been proposed, such as image labeling, image and video description and visual question answering. 
In particular, prominent results have been achieved in image description with various deep neural network architectures, e.g. \cite{lin}, \cite{xu}, \cite{karpathy}, \cite{vinyals}. 
However, the need of generating more narrative texts from images which may reflect experiences, rather than just listing objects and their attributes, has given rise to tasks such as visual storytelling \cite{huang2016}. 
This task is about generating a story from a sequence of images.
Figure~\ref{fig:sisvsdii} shows the difference between descriptions of images in isolation and stories for images in sequence.

In this paper, we describe the deep neural network architecture we used for the Visual Storytelling Challenge 2018. 
The problem to solve in this challenge can be stated as follows: \textit{Given a sequence of 5 images, the system should output a story related to the content and events in the images.}
Our architecture is an extension of the image description architecture presented by~\cite{vinyals}. We submitted the generated stories to the internal track of the Visual Storytelling (VIST) Challenge, which were evaluated using the METEOR metric~\cite{banerjee} as well as human ratings.

\section{Previous work}

The work by~\cite{parkkim} presented probably the first system for generating stories from an album of images. This early approach involved the use of the NYC and Disney datasets mined from blog posts by the authors.

The visual storytelling task and dataset were introduced by~\cite{huang2016}. This was the first dataset specifically created for visual storytelling. They proposed a baseline approach which consists of a sequence to sequence model, where the encoder takes the sequence of images as input and the decoder takes the last state of the encoder as its first state to generate the story. 
Since this model produces stories with generic phrases, they used decode-time heuristics to improve the generated stories. 

\cite{licheng} presented a multi-task model that performs album summarization and story generation. Even though the model achieved state-of-the-art scores on the VIST dataset with different metrics, some of the sample stories presented in the paper are incoherent. 

\section{Model}
\begin{figure}[tb]
    \centering
    \includegraphics[width=.5\textwidth]{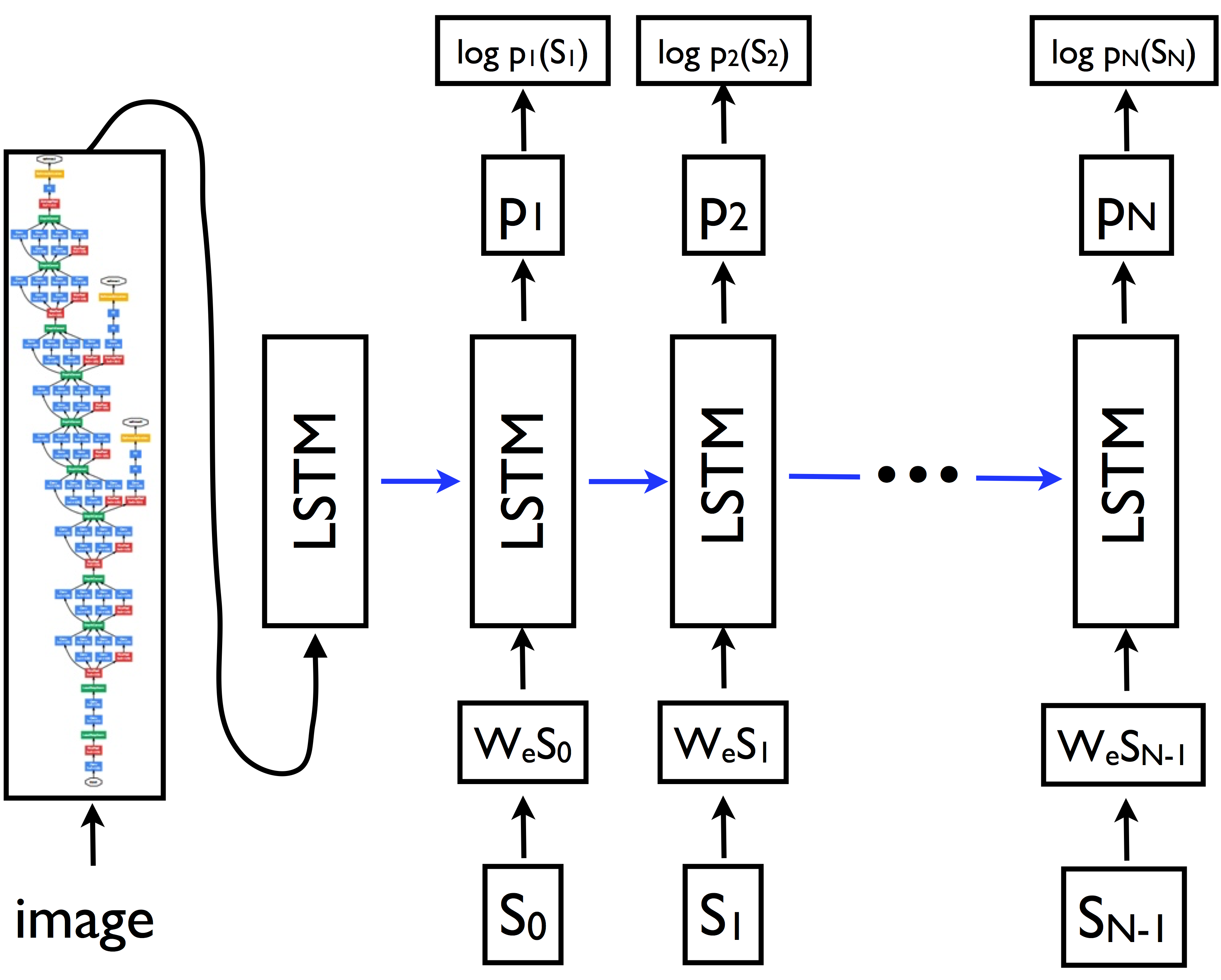}
    \caption{Show and Tell architecture. Image reproduced from~\cite{vinyals}.}
    \label{fig:im2txt}
\end{figure}

Our model extends the image description model by~\cite{vinyals}, which consists of an encoder-decoder architecture. The encoder is a Convolutional Neural Network (CNN) and the decoder is a Long Short-Term Memory (LSTM) network, as presented in Figure \ref{fig:im2txt}. The image is passed through the encoder generating the image representation that is used by the decoder to know the content of the image and generate the description word by word. In the following, we describe how we extended this model for the visual storytelling task.

\subsection{Encoder}

The model's first component is a Recurrent Neural Network (RNN), more precisely an LSTM that summarizes the sequence of images. At every timestep $t$ the network takes as input an image $I_i$ where $i\in\{1,2,3,4,5\}$ from the sequence. At time $t=5$, the LSTM has encoded the 5 images and provides the sequence's context through its last hidden state denoted by $h_e^{(t)}$. The representation of the images was obtained through Inception V3.

\subsection{Decoder}

The decoder is the second LSTM network that uses the information obtained from the encoder to generate the sequence's story.   
The first input $x_0$ to the decoder is the image for which the text is being generated. 
The last hidden state from the encoder $h_e^{(t)}$ is used to initialize the first hidden state of the decoder $h_d^{(0)}$. 
With this strategy, we provide the decoder with the context of the whole sequence and the content of the current image (i.e. global and local information) to generate the corresponding text that will contribute to the overall story. 

Our model contains five independent decoders, one for each image in the sequence. 
All the 5 decoders use the last hidden state of the encoder (i.e. the context) as its first hidden state and take the corresponding image embedding as its first input. 
In this way, the first decoder generates the sequence of words for the first image in the sequence, the second decoder for the second image in the sequence, and so on. 
This allows each decoder to learn a specific language model for each position of the sequence.
For instance, the first decoder will learn the opening sentences of the story while the last decoder the closing sentences.
The word embeddings were computed using word2vec~\cite{mikolov}.

\begin{figure*}[h!]
    \centering
    \includegraphics[trim={0 2cm 0 0},clip, width=\textwidth]{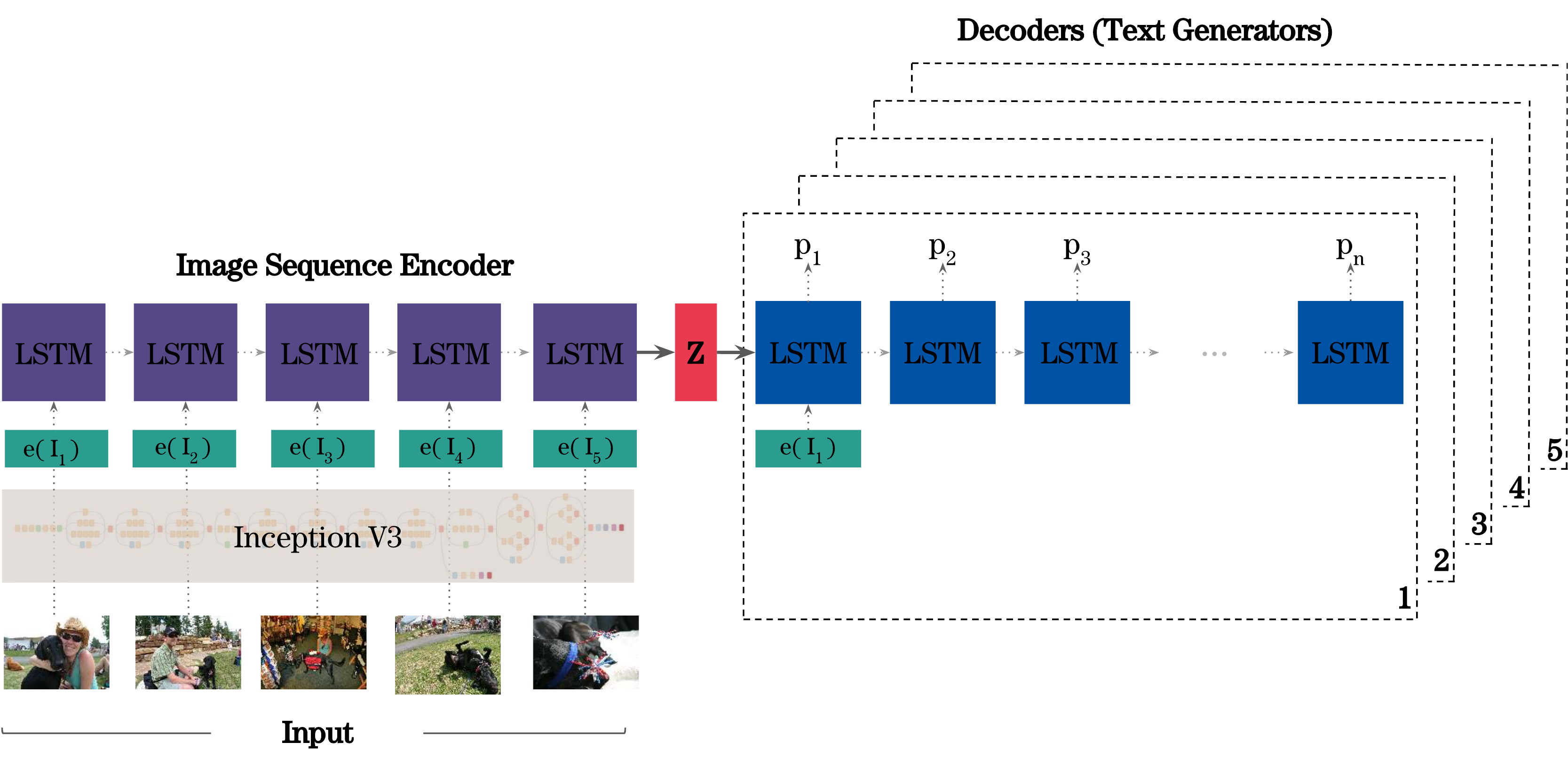}
    \caption{Proposed sequence to sequence architecture.} 
    \label{fig:model}
\end{figure*}

Our proposed architecture is presented in Figure \ref{fig:model}. For each image in the sequence, we obtain its representation $\{e(I_1),...,e(I_5)\}$ using Inception v3. The encoder takes the images in order, one at every timestep $t$. At time $t=5$, we obtain the context vector through $h_e^{(t)}$ (represented by $\mathbf{Z}$). This vector is used to initialize each decoder's hidden state while the first input to each decoder is its corresponding image embedding $e(I_i)$. Each decoder generates a sequence of words $\{p_1,...,p_{n}\}$ for each image in the sequence. The final story is the concatenation of the output of the 5 decoders.

\section{Evaluation}

\subsection{Methodology}
The generated stories were evaluated using both automatic metrics and human ratings.
The automatic evaluation was performed by computing the METEOR metric~\cite{banerjee} on a public test set and a hidden test set. 
The former is a set of $1,938$ image sequences and stories taken from the test set of the VIST dataset~\cite{huang2016}. 
The latter consists of new stories generated by humans from a subset of image sequences of the public test set.
 
 Human ratings of the stories were collected from crowd workers in Amazon Mechanical Turk. 
 Only 200 stories were selected from the hidden test set for this evaluation. 
 The crowd workers evaluated each story on a Likert scale with respect to 6 aspects: \textbf{a)} the story is focused, \textbf{b)} the story has good structure and coherence, \textbf{c)} would you share this story, \textbf{d)} do you think this story was written by a human, \textbf{e)} the story is visually grounded and \textbf{f)} the story is detailed. 
 The crowd workers were also asked to evaluate stories generated by humans for comparison purposes.

\subsection{Results}
Table \ref{table:meteor} shows the METEOR scores by our model in the public and hidden test set of the Visual Storytelling Challenge 2018. 
 Table \ref{table:human} presents results of the human evaluation. Our model achieved competitive METEOR scores in both test sets and performed well in the human evaluation.
 
 \begin{table}[h!]
    \begin{center}
      \begin{tabular}{ | c | c | }
        \hline
         \bf Public Test Set & \bf Hidden Test Set \\ \hline
         .3088 & .3100 \\ \hline
      \end{tabular}
      \caption{Automatic evaluation of stories generated by our visual storyteller using the METEOR metric.}
      \label{table:meteor}
    \end{center}
\end{table}

\begin{table*}[t]
    \begin{center}
      \begin{tabular}{| c | c | c | c | c | c | c | c |}
        \hline
         & \bf a) & \bf b) & \bf c) & \bf d) & \bf e) & \bf f) & \bf Total score \\ \hline
         \bf Ours & 3.347 & 3.278 & 2.871 & 3.222 & 2.886 & 2.893 & 18.498 \\ \hline
         \bf Human & 4.025 & 3.975 & 3.772 & 4.003 & 3.965 & 3.857 & 23.596 \\ \hline
      \end{tabular}
      \caption{Human evaluation of stories generated by our visual storyteller, compared to stories generated by humans.}
      \label{table:human}
    \end{center}
\end{table*}

\begin{table*}[t]
    \begin{center}
      \begin{tabular}{| c | c | c | c | c | c | c | c |}
        \hline
         & \bf BLEU-1 & \bf BLEU-2 & \bf BLEU-3 & \bf BLEU-4 & \bf METEOR & \bf ROUGE & \bf CIDEr  \\ \hline
         \bf Huang et al. & - & - & - & - & 31.4 & - & -\\ \hline
         \bf Yu et al. & - & - & 21.0 & - & 34.1 & \bf 29.5 & \bf 7.5\\ \hline
         \bf Ours & 60.1 & 36.5 & \bf 21.1 & 12.7 & \bf 34.4 & 29.2 & 5.1\\ \hline
      \end{tabular}
      \caption{Automatic evaluation on the VIST dataset. A comparison between the baseline~\cite{huang2016},  \cite{licheng} and ours.}
      \label{table:test}
    \end{center}
\end{table*}

\begin{figure*}[hb!]
    \centering
    \includegraphics[trim={0 4.5cm 0 0},clip,width=\textwidth]{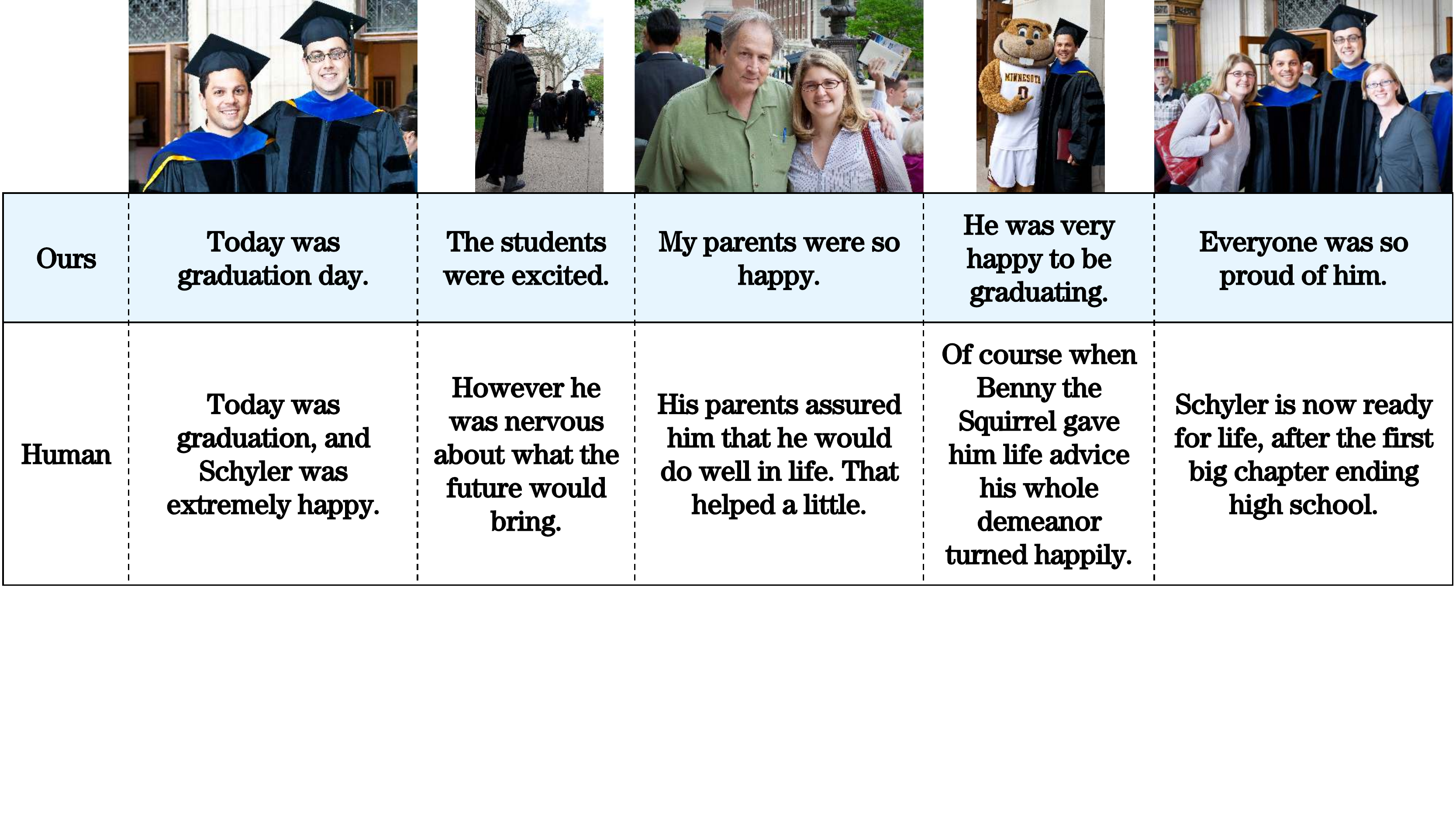}
    \includegraphics[trim={0 4cm 0 0},clip, width=\textwidth]{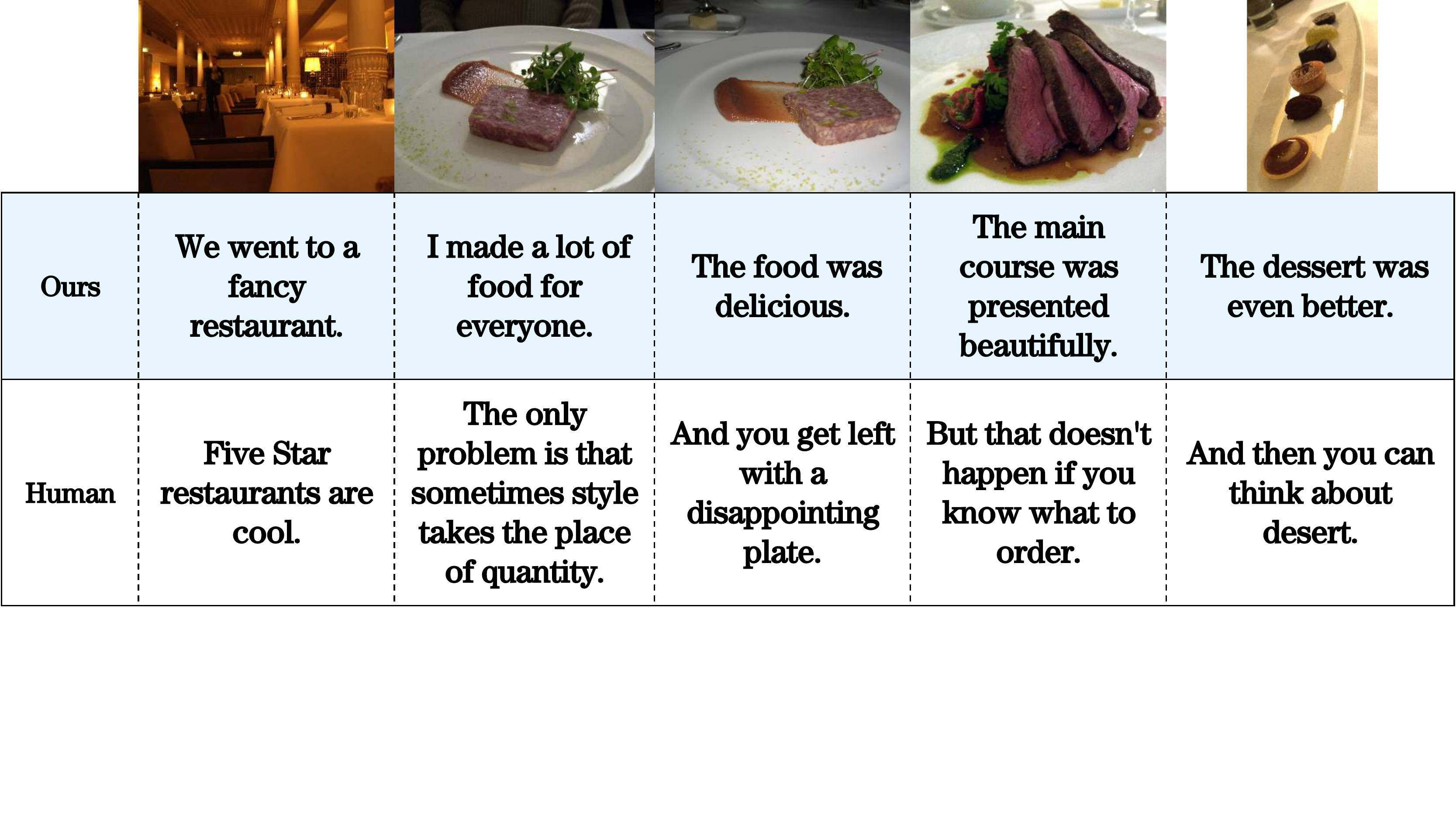}
    \includegraphics[trim={0 7cm 0 0},clip, width=\textwidth]{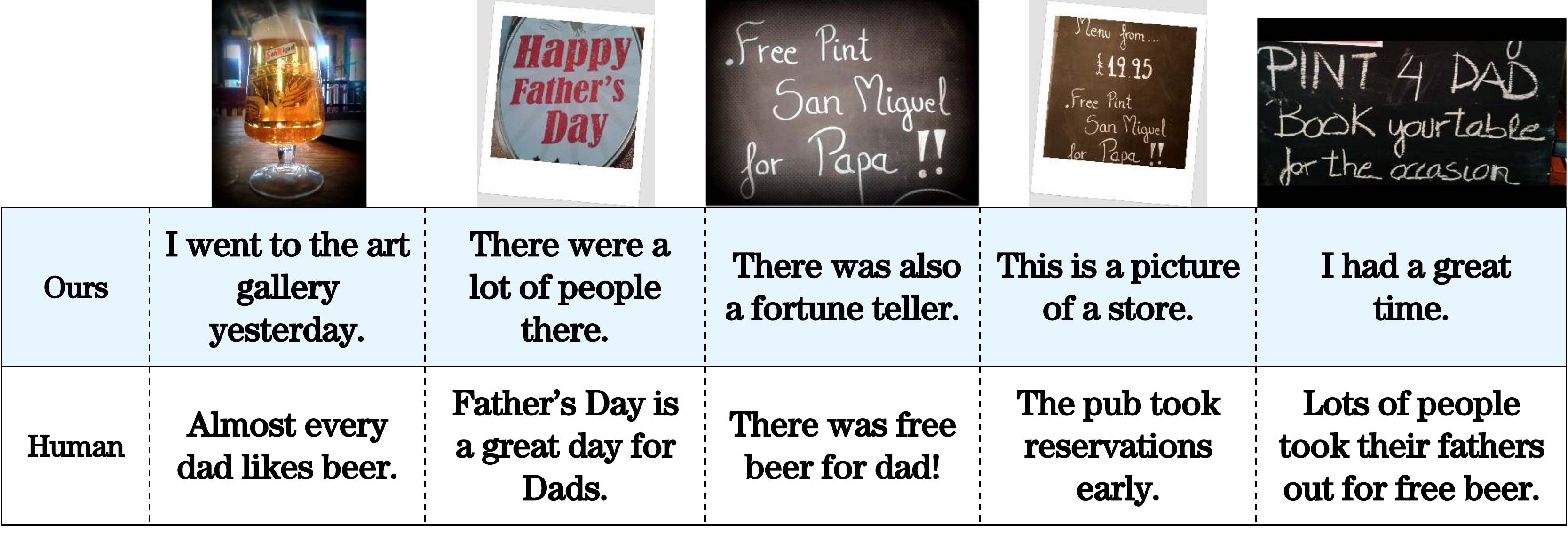}
    \caption{Sample stories generated by our visual storyteller, compared to generated by humans.}
    \label{fig:mesh2}
\end{figure*}
An evaluation over the complete VIST test set was also performed and the results are shown in Table \ref{table:test}\footnote{\footnotesize We used the code available at \href{https://github.com/lichengunc/vist_eval}{github.com/lichengunc}}. 
\newpage
Our model obtained the highest scores with the METEOR and BLEU-3 metrics but lagged behind the model by~\cite{licheng} with the ROUGE and CIDEr metrics. 

Figure \ref{fig:mesh2} shows some sample stories generated by our model from the public test set of the Visual Storytelling Challenge 2018. Although some of the generated stories are grammatically correct and coherent, they tend to contain repetitive phrases or ideas. We can also observe that some stories are not nearly related to the actual content of the images or include generic phrases like \textit{This is a picture of a store}. These limitations of our model reflected on the ratings of the \textbf{visually grounded} and \textbf{detailed} aspects of the human evaluation.

\section{Conclusions and Future Work}
Our visual storyteller incorporates a context encoder and multiple independent decoders to the image description architecture by \cite{vinyals} to generate stories from image sequences. Having an independent decoder for each position of the image sequence, allowed our visual storyteller to build more specific language models using the context vector as its first state and the image embedding as its first input. In the internal track of the Visual Storytelling Challenge 2018, we obtained competitive METEOR scores in both the public and hidden test sets and performed well in the human evaluation. 

In the future, we plan to explore the use of an attention mechanism or a bidirectional LSTM to cope with repetitive phrases within the same story.

\section{Acknowledgments}
This research is supported by the PAPIIT-UNAM research grant IA104016. 
Diana Gonz\'alez Rico is supported by CONACYT.  

\bibliography{naaclhlt2018}
\bibliographystyle{acl_natbib}

\end{document}